\documentclass{article}
\usepackage{ijcai21}

\usepackage{times}

\usepackage{soul}
\usepackage{url}
\usepackage[hidelinks]{hyperref}
\usepackage[utf8]{inputenc}
\usepackage[small]{caption}
\usepackage{graphicx}
\usepackage{amsmath}
\usepackage{amssymb}
\usepackage{booktabs}
\usepackage{bm}
\usepackage{color}
\usepackage{bbm}
\usepackage{footnote}
\makesavenoteenv{table}
\usepackage{varwidth} 

\usepackage{footnote}
\makesavenoteenv{tabular}

\title{UIBert: Learning Generic Multimodal Representations for UI Understanding}

\author{
Chongyang Bai$^1$\thanks{Equal contribution to the work.}\thanks{Work done during internship at Google.}
\and
Xiaoxue Zang$^2$\footnotemark[1]\and
Ying Xu$^{2}$\and
Srinivas Sunkara$^2$\and
Abhinav Rastogi$^2$\and \\
Jindong Chen$^2$\And
Blaise Agüera y Arcas$^2$\\
\affiliations
$^1$Dartmouth College\\
$^2$Google Research\\
\emails
\ bchy1023@gmail.com,\\
\{xiaoxuez,yingyingxuxu,srinivasksun,abhirast,jdchen,blaisea\}@google.com
}

\newcommand{\uibert}{{\texttt{UIBert}}}

\begin{document}

\maketitle

\begin{abstract}
To improve the accessibility of smart devices and to simplify their usage, building models which understand user interfaces (UIs) and assist users to complete their tasks is critical. However, unique challenges are proposed by UI-specific characteristics, such as how to effectively leverage multimodal UI features that involve image, text, and structural metadata and how to achieve good performance when high-quality labeled data is unavailable. To address such challenges we introduce UIBert, a transformer-based joint image-text model trained through novel pre-training tasks on large-scale unlabeled UI data to learn generic feature representations for a UI and its components.
Our key intuition is that the heterogeneous features in a UI are self-aligned, i.e., the image and text features of UI components, are predictive of each other. We propose five pretraining tasks utilizing this self-alignment among different features of a UI component and across various components in the same UI. We evaluate our method on nine real-world downstream UI tasks where UIBert outperforms strong multimodal baselines by up to 9.26\% accuracy. 
\end{abstract}

\section{Introduction}
As an increasing number of people rely on smart devices to complete their daily tasks, user interface (UI) - the tangible media through which human interacts with the various applications, plays an important role in creating a pleasant user interaction experience. Recently, many UI related tasks have been proposed to improve device accessibilities and assist device operations. For instance, ~\cite{li2020mapping} studied how to ground natural language commands (e.g. ``play next song") to executable actions in UIs, which enables voice control of devices for visual or situational (e.g. driving) impaired users. ~\cite{huang2019swire} proposed generating UI descrptions which is useful for screen readers like Talkback\footnote{\url{https://support.google.com/accessibility/android/answer/6283677}}. Some other tasks aim to help UI designers learn best design practices, e.g. retrieving similar UIs~\cite{huang2019swire} or UI elements ~\cite{he2021ab}.

\begin{figure}[t]
\centering
\includegraphics[width=\columnwidth]{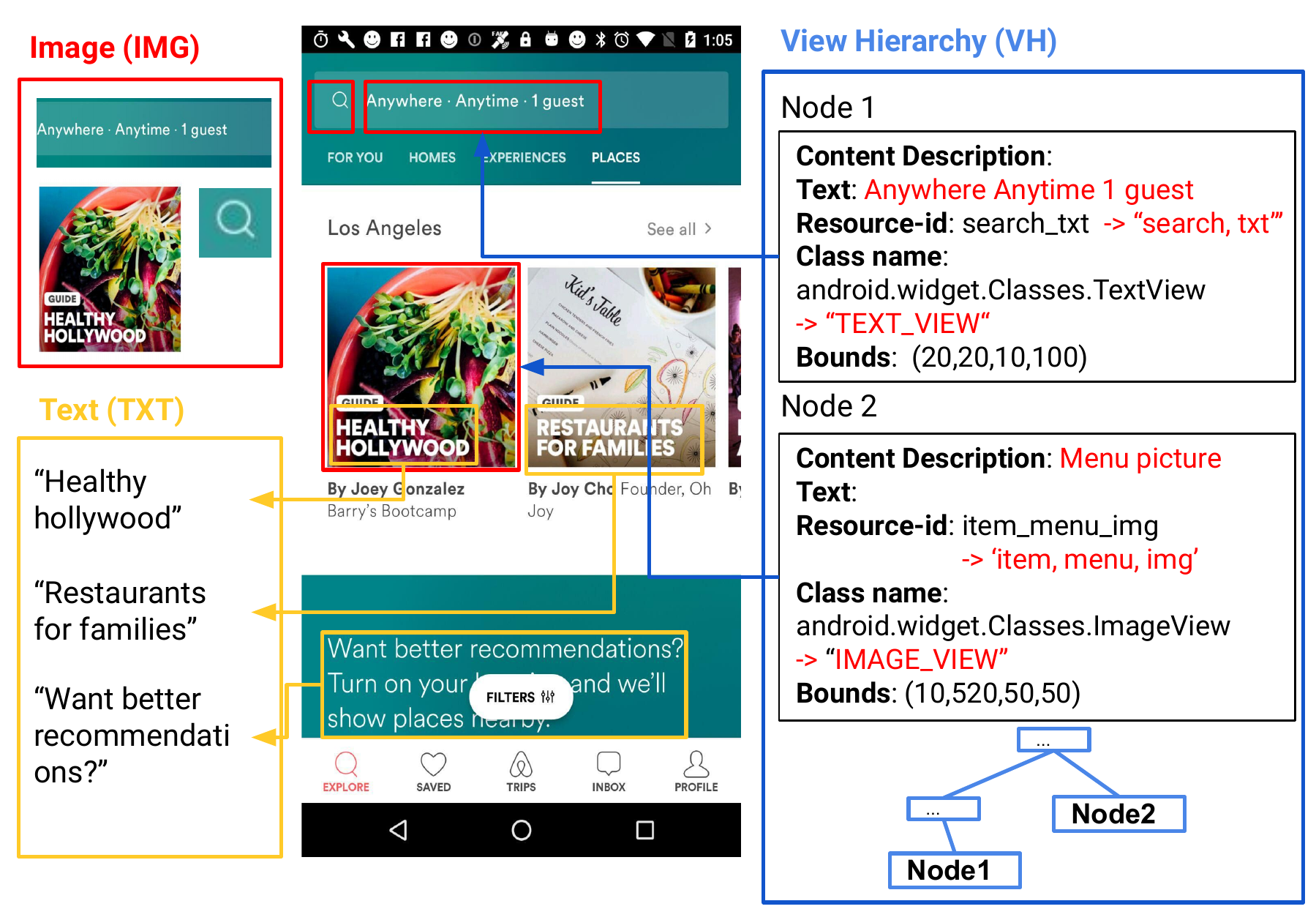}
\caption{Heterogeneous features in a UI.}
\label{fig:inputs}
\end{figure}

All of the above tasks require a comprehensive understanding of the UI, which proposes unique challenges. The first is how to effectively leverage cross-modal knowledge. UI consists of heterogeneous information (Fig. \ref{fig:inputs}) such as images, natural language (e.g. texts on the UI), and structural metadata (e.g. Android view hierarchy in mobile apps and Document Object Model in webpages). Especially, the metadata contains rich information about UI layouts and potentially functionality of UI elements that are invisible to the users, yet also suffering from noise~\cite{li2020mapping}. Previous work usually utilized single-modality data, e.g. only image, to solve the tasks~\cite{liu2018learning,chen2020unblind}. How to effectively leverage cross-modal knowledge and diminish the affect of noise for general UI understanding remains an open question. Second, high-quality task-specific UI data is expensive to achieve as it requires complicated setups of app/web crawlers and time-consuming human labeling work~\cite{li2020widget,swearngin2019modeling}, which inevitably slows the model development cycle. When large-scale data is unavailable, it's non-trivial to overcome overfitting and achieve satisfying performance. 


\begin{figure*}[t]
\centering
\includegraphics[width=\linewidth]{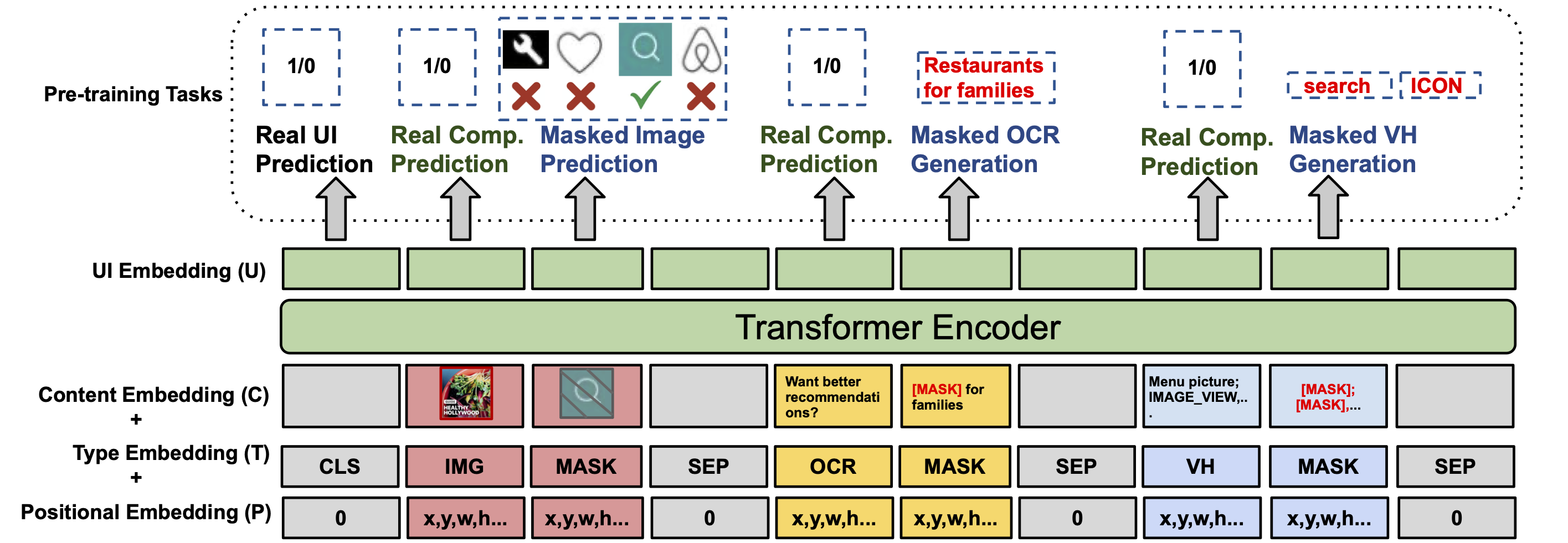}
\caption{\uibert\ overview. It takes Fig~\ref{fig:inputs} as the input. Content, type, and positional embeddings are computed and summed as the input to the Transformer. The output UI embeddings $U$, is used for pre-training and downstream tasks. Note that we randomly choose one type of components (IMG, OCR, or VH) to mask in pretraining but to save space, the figure shows the case when we mask all three types in one UI. \label{fig:uibert_archi}}
\end{figure*}

Inspired by the recent success of self-supervised learning like BERT \cite{devlin2018bert} and its multimodal variants \cite{su2019vl,li2020unicoder}, \cite{he2021ab} explored building generic feature representations for UI from unlabelled data that can be applied to various UI related tasks. 
Their promising results open up a new-emerging research direction and leave ample space of exploration. 
As a concurrent work, we also propose a novel transformer-based multimodal approach \uibert\ (Fig.~\ref{fig:uibert_archi}) that generates contextual UI representations for solving the aforementioned challenges. But different from \citeauthor{he2021ab} that leverages temporal connectivity of UIs in a UI sequence connected by user actions, we utilize the inter-connectivity between heterogeneous features on a single UI. Specifically, our key intuition is that heterogeneous features on a UI are predicative of each other. For example, in Fig.~\ref{fig:inputs} that presents a UI with its 
multimodal features, the texts on the UI (``Healthy hollywood", ``Restaurants for families"), the carousel images about food and menu, and the content description of Node 2 in the view hierarchy (``Menu picture") are all semantically related and indicate the theme of this UI. Based on it, we design five novel pretraining tasks to leverage the alignment between various UI features. We experimentally show that our approach outperforms the prior work on all the downstream evaluation tasks. 
Overall, our contributions are:

\begin{itemize}
\item We propose \uibert\ with five novel pretraining tasks, utilizing the image-text correspondence to learn contextual UI embeddings from unlabeled data. 
\item We evaluate \uibert\ on nine downstream tasks of five categories, including zero-shot evaluations. \uibert\ outperforms strong baselines in all tasks. Qualitative evaluations also proves its effectiveness.
\item We release two new datasets extended from Rico \cite{Deka2017Rico} for two tasks: similar UI component retrieval and referring expression component retrieval. \footnote{https://github.com/google-research-datasets/uibert}
\end{itemize}

\vspace{-1mm}
\section{Related Work}
\vspace{-1mm}
Different machine learning models have been proposed to understand UI. For example, 
\cite{li2020mapping} leveraged Transformer to map natural language commands to executable actions in a UI. \cite{li2020widget,chen2020unblind} used Transformer to generate textual descriptions for UI elements. There were also attempts using convolutional neural networks to retrieve similar UIs for design mining~\cite{Deka2017Rico,liu2018learning,huang2019swire}. Past work generally built task-specific models and required substantial labeled data. In contrast, we focus on learning general knowledge of UI that is applicable for various tasks and leverage large-scale unlabeled data. ActionBert \cite{he2021ab} is the most relevant work to us. They proposed training a Transformer that takes the multimodal features generated by separate image and text encoders through well-designed pre-training tasks. The main difference is that they leveraged the temporal connections between UIs in a UI sequence to design their pretraining tasks while we focus on the self-alignment among different multimodal features in a single UI. 
Additionally, ActionBert freezes the image and text encoders during pretraining, whereas we use trainable lightweight encoders such as Albert \cite{lan2019albert} and EfficientNet \cite{tan2019efficientnet}. This enables representation of domain-specific knowledge within encoder parameters. 


\section{Background}\label{sec:background}
In this section, we introduce the Android view hierarchy which is one of our model inputs, and summarize the original BERT model from which \uibert\ is inspired.

\paragraph{View hierarchy.} View hierarchy is a tree representation of the UI elements created by Android developers\footnote{\url{https://developer.android.com/reference/android/view/View}}. 
Each node describes certain attributes (e.g. bounding box positions, functions) of a UI element - the basic building block of UI.
An example of a view hierarchy tree can be found on the right of Fig.~\ref{fig:inputs}. \emph{Text} records the visible text of textual elements on the screen; \emph{Content description} and \emph{Resource-id} sometimes contain useful information about the functionality (e.g. navigation, share) which are usually invisible to users. \emph{Class name} is the categorical Android API class name defined by developers, and \emph{Bounds} denotes the element's bounding box location on the screen. Note that except for \emph{Class name} and \emph{Bounds}, the other fields can be empty. Although view hierarchy is informative, it is noisy~\cite{li2020mapping} and is not completely standardized that different view hierarchies can lead to the same screen layout. Therefore, it alone is insufficient to provide a whole image of the UI. 

\paragraph{BERT.} BERT is a Transformer \cite{vaswani2017attention} based language representation model, which takes as input a sequence of word piece tokens pre-pended with a special [CLS] token. BERT defines two pretraining tasks: 
Masked language model (MLM) that learns the word-level embeddings by inferring randomly masked tokens from the unmasked ones, and next sentence prediction (NSP) that learns the sentence-level [CLS] embedding by predicting if two input sentences are consecutive. 
Inspired by it, \uibert \ adapts MLM to three and NSP to two novel pretraining tasks to learn generic and contextualized UI representations.

\section{Pre-training}
We introduce the details of \uibert\ starting with its multimodal inputs, then the entire architecture, followed by our proposing pretraining tasks, and lastly qualitative evaluations of the pretrained embeddings.
\subsection{Inputs to UIBert}\label{sec:inputs}
 Given a UI image with its view hierarchy, we first obtain three types of UI components: images (IMG), OCR texts (OCR), and view hierarchy nodes (VH) as shown in Fig.~\ref{fig:inputs}. Below illustrates their definitions and the individual component features we extract, which will be used in the next subsection:

\paragraph{VH components.} VH components are leaf nodes\footnote{Other nodes are discarded as they usually describe a collection of UI elements.} of a view hierarchy tree.
For each leaf node, we encode the content of its textual fields - \emph{Text}, \emph{Content description}, \emph{Resource-id}, and \emph{Class name} that are described in Section \ref{sec:background} into feature vectors. As a preprocessing step, we normalize the content of \emph{Class name} by heuristics to one of the 22 classes (e.g. \textit{TEXT\_VIEW}, \textit{IMAGE\_VIEW}, \textit{CHECK\_BOX}, \textit{SWITCH}) and split content of \emph{resource-id} by underscores and camel cases. Normalized \emph{Class name} is then encoded as a one-hot embedding, while the content of other fields are respectively fed into a pretrained Albert \cite{lan2019albert} to obtain their sentence-level embeddings. All the obtained embeddings are concatenated as the final component feature of the VH component.

\paragraph{IMG components.} IMG components are image patches cropped from the UI based on the bounding boxes denoted in the VH components. We use EfficientNet \cite{tan2019efficientnet} of which the last layer is replaced by spatial average pooling to get the component feature of each IMG component.

\paragraph{OCR Components.} OCR components are texts detected by a pretrained OCR model \cite{mlkit-ocr}
on the UI image, which is in most cases complementary with the content in the VH components. We generate its component features using the same Albert model as is used for encoding the VH components.



\subsection{UIBert Architecture}
Fig. \ref{fig:uibert_archi} shows an overview of our model. It takes the aforementioned components (IMG, VH, OCR) in a single UI as input and uses a six-layer Transformer with 512 hidden units and 16 self-attention heads to fuse features of different modalities. Following BERT, we organize the input as: CLS, IMGs, SEP, OCRs, SEP, VHs, SEP, where CLS aims to learn the UI-level embedding and SEP is used to separate UI components of different types. Below describes three kinds of embeddings we compute for \uibert.


\paragraph{Type embedding.} To distinguish input components of diverse types, we introduce six type tokens: IMG, OCR, VH, CLS, SEP, and MASK. MASK is a special type used for pretraining which is discussed in the next subsection. A one-hot encoding followed by linear projection is used to get the type embedding, $T_i \in \mathbb{R}^d$, for the $i_{th}$ component in the sequence where $d$ is the dimension size that is 512 in our case. 
\paragraph{Positional embedding.} We encode the location
feature of each component using its bounding box, which consists of normalized top-left, bottom-right point coordinates, width, height, and area of the bounding box. Similar to type embeddings, a linear layer is used to project the location feature to the positional embedding, $P_i \in \mathbb{R}^d$, for the $i_{th}$ component ($P_i=\bm{0}$ for CLS and SEP).

\paragraph{Content embedding.} We linearly project the extracted component features (Sec. \ref{sec:inputs}) to the content embedding $C_i \in \mathbb{R}^d$, for every $i_{th}$ input with $type(i) \in \{\text{IMG, OCR, VH}\}$ and use 0s for the inputs of other types.

The final input to the Transformer is constructed by summing all the above three embeddings, and \uibert\ generates the final UI embeddings $U \in \mathbb{R}^{n \times d}$ by:
\begin{equation}
U = \text{TransformerEncoder}(T+P+C),
\end{equation}
where $T, P, C \in\mathbb{R}^{n \times d}$ and $n$ is the sequence length.


\subsection{Pre-training Tasks}
We design five novel pre-training tasks. The first two aim to learn the alignment between UI components of different types (e.g. VH and IMG) by creating unaligned fake UIs and training the model to distinguish them from real ones. The last three are inspired by the MLM task in BERT: for each UI, we choose a single type (IMG, OCR or VH), randomly mask 15\% of the UI components of that type , and infer their content from the unmasked ones. Our pretraining dataset consists of 537k pairs of UI screenshots and their view hierarchies obtained using the Firebase Robo app crawler \cite{robodocs}. We use Adam \cite{kingma2014adam} with learning rate 1e-5, $\beta_1=0.9, \beta_2=0.999, \epsilon=$1e-7 and batch size 128 on 16 TPUs for 350k steps. The five tasks are defined below.

\begin{figure}
\includegraphics[width=\columnwidth]{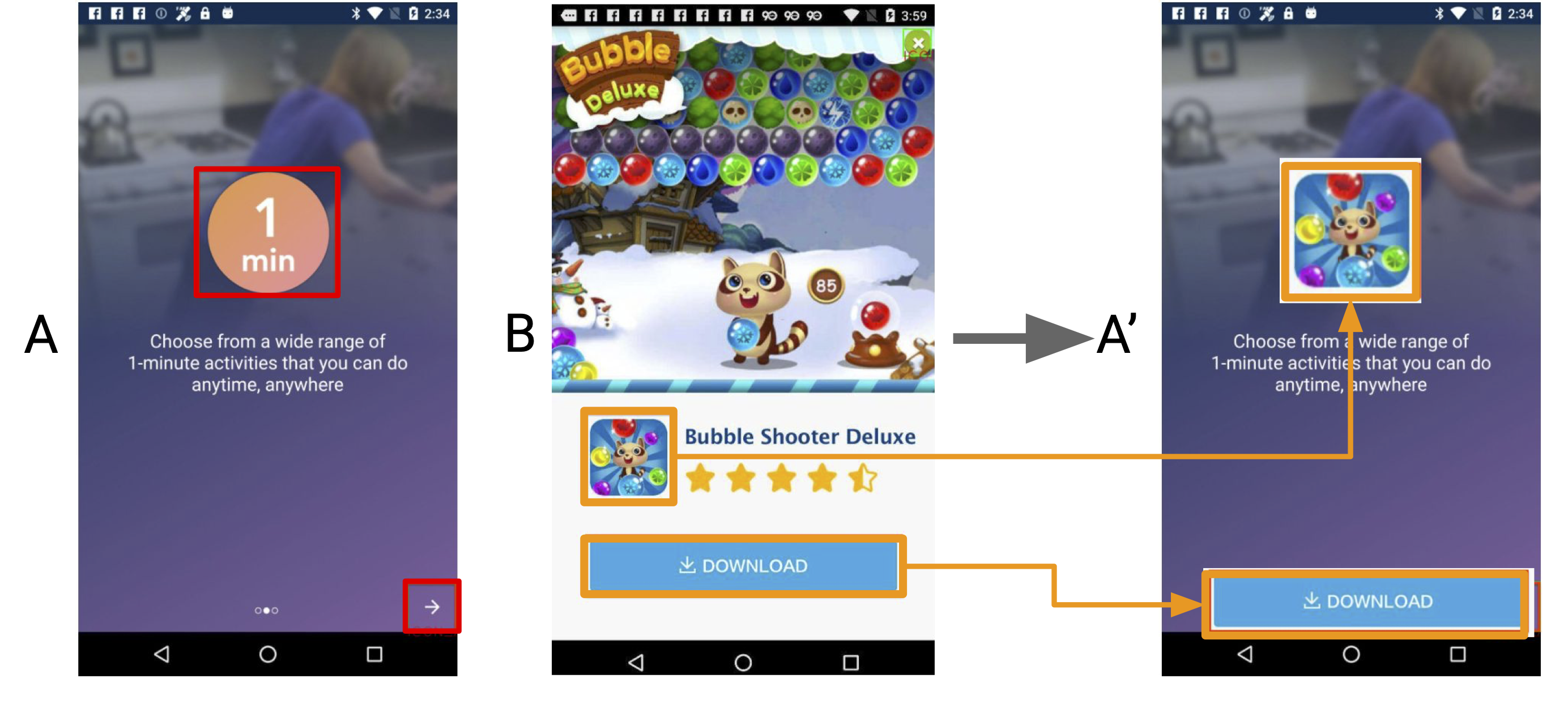}
\centering
\caption{Fake UI generation. 15\% of the components in UI-A are randomly chosen (red boxes) and replaced by the same amount of random components in UI B (yellow boxes) to get A'.\label{fig:fake_ui}}
\end{figure}

\paragraph{Task 1: Real UI Prediction (RUI).} Given an original UI-A, we create a fake version of it, A', by replacing 15\% of its UI components with components from UI-B, which is randomly selected from UIs in the same batch. For each UI, initially the type of components to replace is randomly chosen (IMG, OCR or VH). An example is shown in Fig. \ref{fig:fake_ui}, where two IMG components in UI-A are replaced by two IMG components from UI-B to yield the fake UI-A'. 
Note that in this case, we do not change the VH and OCR inputs to the Transformer as we try to make the task harder by having only small difference between the original and fake UI.
The RUI task predicts whether a UI is real or not by minimizing the cross-entropy  (CE) objective:
\begin{equation}
L_{RUI} = CE(y, \hat{y}),
\label{equ:l_rui}
\end{equation}
where $y$ is the binary label for UI $\bm{x}$ ($y=1$ if $\bm{x}$ is real), and $\hat{y}=\text{Sigmoid}(FC(U_{CLS}))$ is the prediction probability. $U_{CLS}$ corresponds to the output embedding of CLS token (Fig. \ref{fig:uibert_archi}), and $FC$ is a fully connected layer.

\paragraph{Task 2: Real Component Prediction (RCP).} 
We further predict for every fake UI, whether a UI component aligns with the rest or not. In UI-A' of Fig. \ref{fig:fake_ui}, only the two IMG components that are switched from UI-B are fake, whereas all OCR and VH components and the rest of IMG components are real. Intuitively, the content of a fake component would not align with the rest of the components and the model needs to learn from the context to make the correct prediction.
The objective of RCP is the sum of the weighted cross-entropy loss over all UI components in a \emph{fake} UI:

\begin{equation}
L_{RCP} =  \sum_{type(i) \in \{\text{IMG, OCR, VH}\}} CE(y_i, \hat{y_i}; \lambda),
\end{equation}
where $y_i$ is the label of the $i_{th}$ component, and $\hat{y_i}$ is the prediction made by a linear layer connected to the UI embedding $U_i$. The weight $\lambda$ is multiplied to the loss for fake components to address the label imbalance. We use $\lambda=2$ in our case.

\paragraph{Task 3: Masked Image Prediction (MIP).} 
We randomly mask 15\% of the IMG inputs by replacing its content embeddings with 0s and its type feature with MASK. This task aims to infer the masked IMG inputs from its surrounding inputs for each real UI. Prior work on multimodal pretraining also designed similar tasks, but most of them try to predict either the object class (e.g. tree, sky, car) \cite{li2020unicoder} or the object features \cite{su2019vl} of the masked image patches, which are obtained by a pre-trained object detector. However, such methods highly rely on the accuracy of the pretrained object detector and is unsuitable for our case, as there is no existing object detector specifically trained with UI data to detect all the generic UI components. Thus, we try to predict the masked IMG inputs in a contrastive learning manner (Fig. \ref{fig:uibert_archi}): 
given the content embedding of the original IMG component (positive) with the content embeddings of some negative distracting IMG components sampled from the same UI, the output embedding of the masked positive is expected to be closest to its content embedding in terms of their cosine similarity scores. 
Formally, let $\mathcal{M}_{\text{IMG}}$ be the set of masked IMG indices in a \emph{real} UI. We employ the softmax version of Noise Contrastive Estimation (NCE) loss \cite{jozefowicz2016exploring} as the objective:

\begin{equation}
L_{MIP} = - \sum_{i \in \mathcal{M}_{\text{IMG}}} \log NCE(i | \mathcal{N}(i)),
\label{equ:l_mop}
\end{equation}
\begin{equation}
NCE(i | \mathcal{N}(i)) = \frac{\exp(U_i^T C_i)}{\exp(U_i^T C_i) + \sum_{j \in \mathcal{N}(i)}\exp(U_i^T C_j)},
\end{equation}
where $\mathcal{N}(i)$ is the set of negative IMG components for $i$. In practice, we use the $k$ closest IMGs to the masked component $i$ in the image as the negative components.

\paragraph{Task 4: Masked OCR Generation (MOG).} When masking OCR inputs, as each OCR component is a sequence of words, we frame the prediction of the masked OCR as a generation problem -- a 1-layer GRU decoder \cite{chung2014empirical} takes the UI embedding of the masked OCR component as input to generate the original OCR texts. We use a simple decoder as our goal is to learn powerful UI embeddings. Since it can be hard to generate the whole sequence from scratch, we mask tokens of a masked OCR component with probability of 15\% (e.g. only "Restaurants" is masked in the OCR component "Restaurants for families" in Fig. \ref{fig:uibert_archi}). Denote $t_i = (t_{i,1}, \ldots t_{i,n_i})$ as the WordPiece \cite{wu2016google} tokens of OCR component $i$ where $t_{i,j}, \forall j$ is the one-hot encoding of the $j$th token, and $\hat{t}_i=GRU(U_i)=(\hat{t}_{i,1}, \ldots \hat{t}_{i,n_i})$ as the predicted probability of the generated tokens, the MOG objective is framed as the sum of multi-class cross-entropy losses between the masked tokens and generated ones:
\begin{equation}
L_{MOG} = \sum_{(i, j) \in \mathcal{M}_{\text{OCR}}} CE(t_{i,j}, \hat{t}_{i,j}),
\label{equ:l_mog}
\end{equation}
where $\mathcal{M}_{\text{OCR}}$ denotes the set of (compomnent id, token id) pairs of the masked OCRs. 

\paragraph{Task 5: Masked VH Generation (MVG).} 
For VH components, we observe that \emph{Resource-id} is usually short that contains only two to three tokens and \emph{Text} field overlaps with OCR texts. Hence, we only mask the \emph{Content description} and \emph{Class name}. For each masked VH component, we generate its \emph{Content description} using the same GRU decoder as for the MOG task, and predict the \emph{Class name} label by a fully connected layer with a softmax activation.
Formally, 
\begin{equation}
L_{MVG} = \sum_{i \in \mathcal{M}_{\text{VH}}} (CE(c_i, \hat{c}_i) + \sum_j CE(t_{i,j}, \hat{t}_{i,j})),
\end{equation}\label{equ:l_mvg}
where $\mathcal{M}_{\text{VH}}$ is the set of masked VH components, $c_i$ is the one-hot encoding of the \emph{Class name} label of VH component $i$, $\hat{c}_i=\text{Softmax}(FC(U_i))$ is the predicted probability vector, $t_{i,j}, \hat{t}_{i,j}$ represent the original and predicted content description tokens following the same definition as the OCR tokens.

In practice, content descriptions of UI components can be used by screen reading tools to provide hints for people with vision impairments, yet prior work shows that more than 66\% buttons are missing content description \cite{chen2020unblind}. We show in Sec. \ref{sec:quality_eval} that \uibert\ pre-trained with the MVG task can generate meaningful descriptions and has great potential to assist screen readers.

Overall, the pretraining loss objective for a UI is
\begin{equation}
\small
L = L_{RUI} + \mathbf{1}_{\{y=0\}} L_{RCP} + \mathbf{1}_{\{y=1\}} (L_{MIP} + L_{MOG} + L_{MVG}), 
\label{equ:l_all}
\end{equation}
where $\mathbf{1}_{\{.\}}$ is the indicator function. 

\subsection{Qualitative Evaluation}\label{sec:quality_eval}
To verify the effectiveness of our pretraining tasks, we visualize the UI embeddings of a pre-trained \uibert, and showcase the generated content descriptions without any fine-tuning. 

\paragraph{Embedding visualization.}

We use t-SNE \cite{maaten2008visualizing} to visualize $U_{CLS}$ of the UIs that belong to the top-5 common app types and the embeddings of UI components of the top-5 common icon types in Rico~\cite{Deka2017Rico}, which is a public mobile design dataset containing 72k unique UI screenshots with view hierarchies crawled across 9.7k mobile apps. 
We observe that embeddings of the same app types or icon types are grouped together, suggesting that \uibert\ captures meaningful UI-level and component-level features.

\begin{figure}
\centering
  \includegraphics[width=\linewidth]{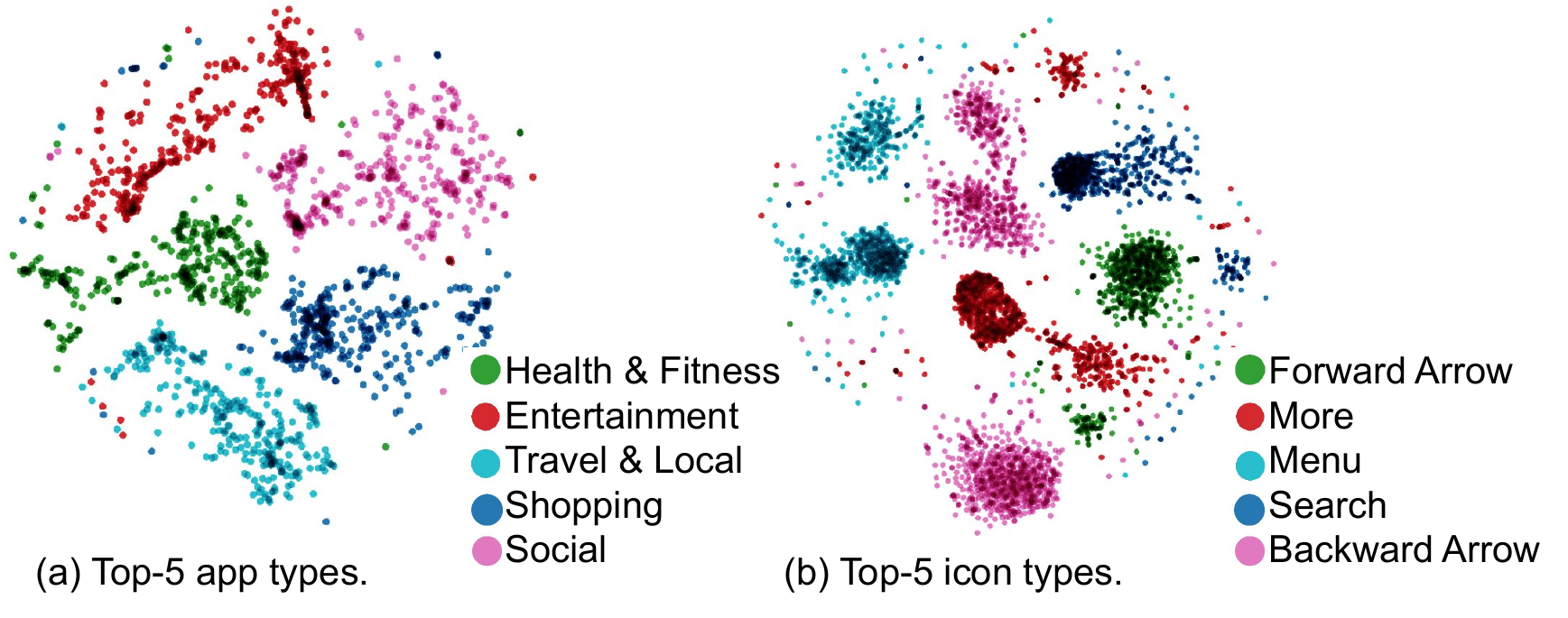}
\caption{Zero-shot embedding visualization of UIs of top-5 common apps and UI components of top-5 common icon types using t-SNE. Best viewed in color.\label{fig:emb_tsne}}
\end{figure}

\paragraph{Content description generation.}

\begin{figure}[ht]
\centering
\includegraphics[width=\linewidth]{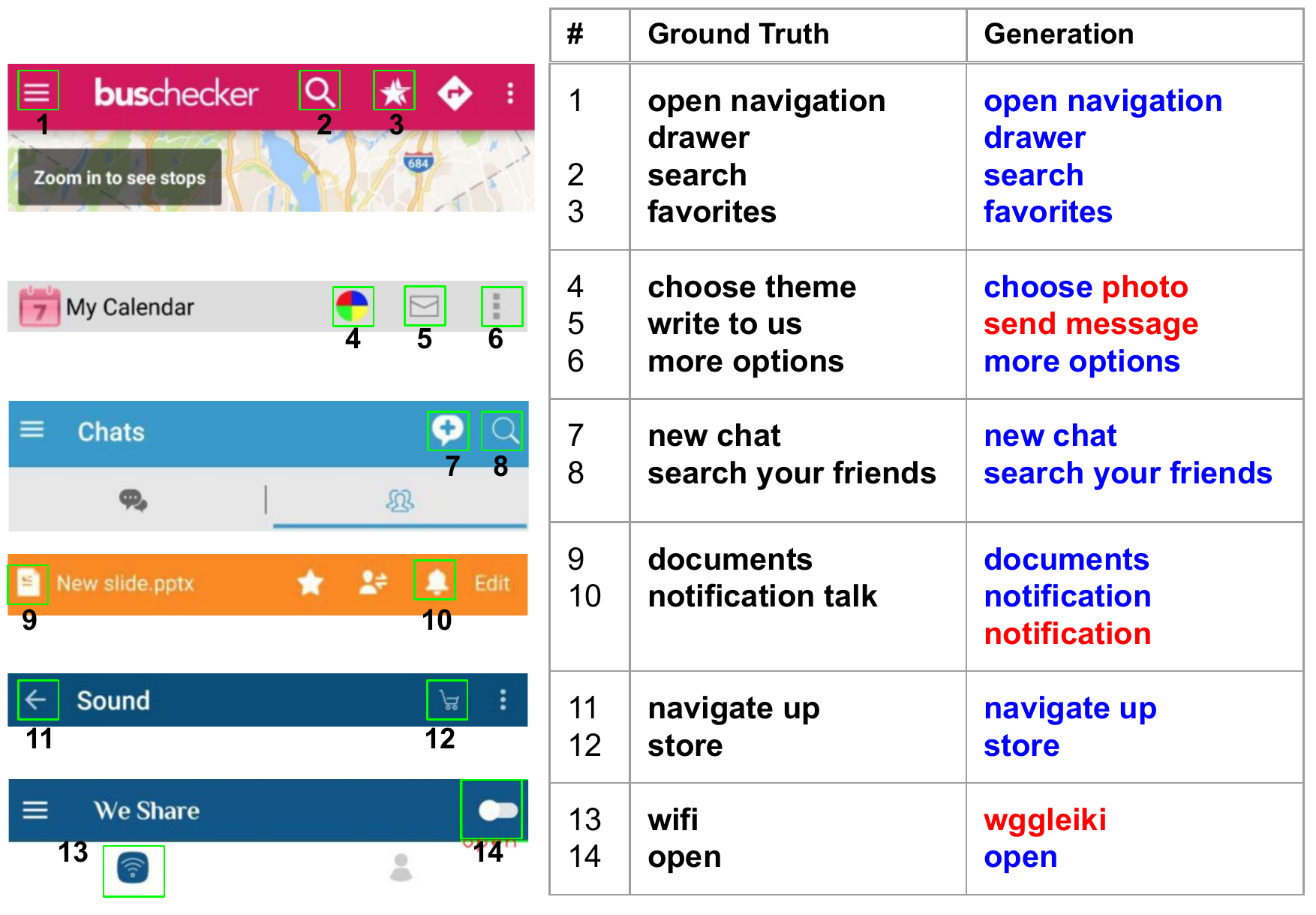}
\caption{Examples of the generated content descriptions by a pretrained \uibert\ (correct in blue and incorrect in red).\label{fig:content_desp}}
\end{figure}

We generate content descriptions for the synchronized UIs in the public RicoSCA dataset \cite{li2020mapping} (details in Sec. \ref{sec:img-sync-cls}).
We mask all the content descriptions in the input and generate them following the same settings in the MVG task. As shown in examples of Fig. \ref{fig:content_desp},
most of the generated descriptions are correct. Some are incorrect but reasonable (case 5). Overall, 70\% of the generated content descriptions are the same as the ground truth. 

\section{Downstream Tasks}
As \uibert\ is designed to learn generic contextual UI representations transferable to various UI understanding tasks, we also conduct experiments to evaluate its performance on downstream tasks. We choose nine practical downstream tasks across five categories, including two zero-shot tasks. Our finetuning approach introduces minimal 
task-specific parameters and finetunes all the parameters end-to-end. For each finetuning task, we train the model for 200k steps with dropout rate of 0.1, and use the same optimizer configuration and batch size as that in pretraining. In the following, we first describe the baselines to compare with, then the details of each downstream task including definition, datasets, experimental setups and results. 

\subsection{Baselines} 
We consider two baseline encoding mechanisms for the downstream tasks: \emph{EfficientNet+Albert} and \emph{ActionBert}. The first one uses EfficientNet-B0 \cite{tan2019efficientnet} and Albert \cite{lan2019albert} to encode the image and text components of the UI separately. The obtained embeddings are then concatenated and fed into the same prediction head as used in \uibert\ for downstream tasks.
As there is no attention across the two modalities, it serves as an ablation evaluation for the Transformer blocks used in \uibert\ which facilitate this cross-modal attention.
The second baseline, ActionBert \cite{he2021ab}\footnote{We use ActionBert$_{BASE}$ due to its comparable size to \uibert.}, is a recently proposed UI representation model, pretrained with user interaction traces.

\subsection{Similar UI Component Retrieval}
\begin{figure}[]
\includegraphics[width=\columnwidth]{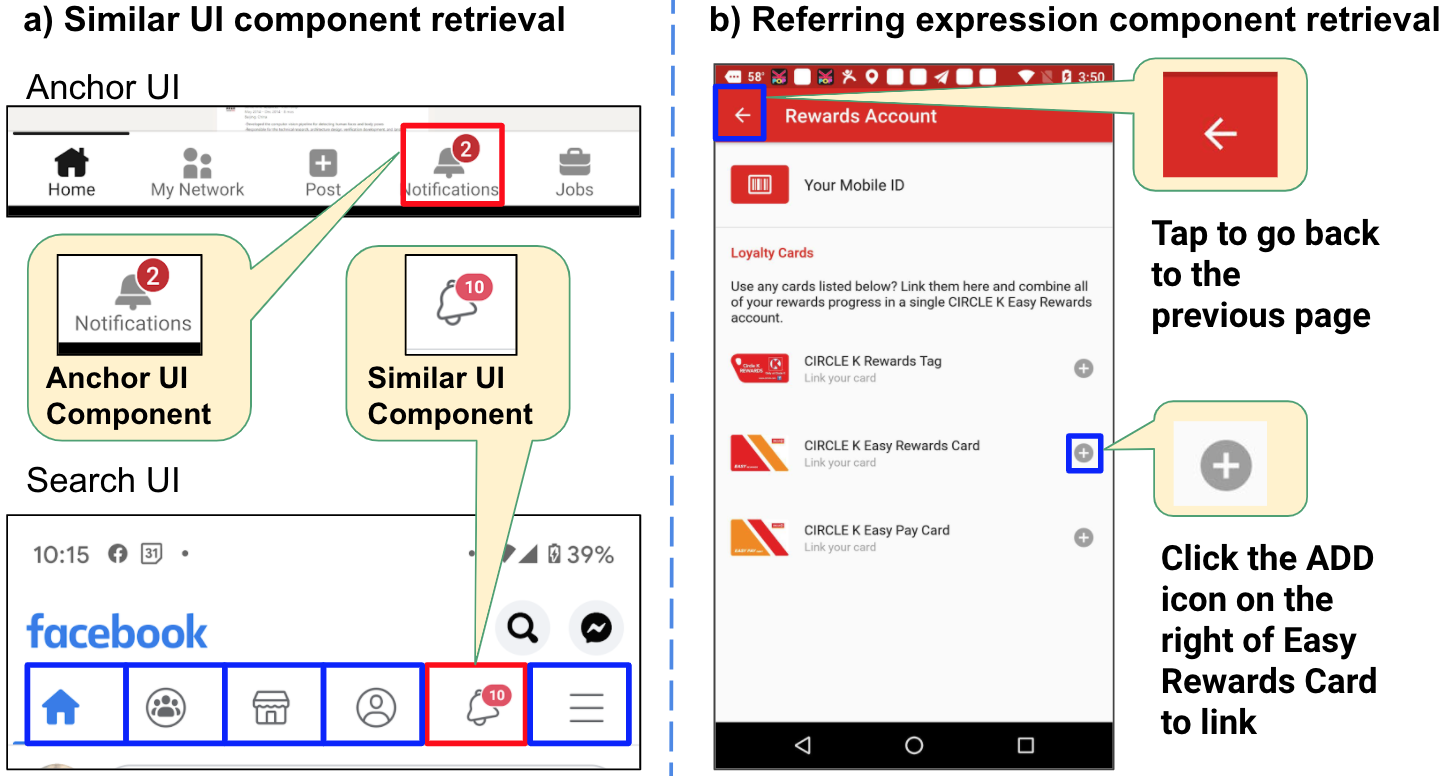}
\centering
\caption{a) An example of similar UI component retrieval. b) An example of referring expression component retrieval.}\label{fig:task-vis}
\end{figure}

In this task, given an anchor UI with an anchor component as query and a seacrh UI with a set of candidate components, the goal is to select the closest candidate to the anchor component in terms of the functionality (Fig.~\ref{fig:task-vis}(a)). Models for this task can assist UI designers to find best design practices. For example, upon creating a new UI, the
designer can refine any component by retrieving similar ones
from a UI database.

One dataset for this task is extended from Rico \cite{Deka2017Rico} which serves as a database of mobile app UIs. It consists of 1M anchor-search UI component pairs annotated via crowd-sourcing and programmatic rules. We use 900k pairs for training, 32k pairs for dev, and 32k pairs for test. 
On average, each search UI has 10 candidate components for the model to choose from. 
Another dataset for this task includes 65k anchor-search web UI pairs. Each search web UI has 35 components on average. Note that as view hierarchies are unavailable in web UIs, there is no VH component input to \uibert\ during finetuning.

To adapt \uibert\ to this task, the anchor UI and search UI are fed into \uibert\ separately to get the output embeddings, then the dot products between embeddings of anchor component and candidate components are used as similarity scores to select the most similar candidate to the anchor. We finetune \uibert\ using the multi-class cross entropy loss on the similarity scores. Since no additional model parameters are needed, the task is also evaluated in a zero-shot manner by directly using the pretrained model. To adapt the \emph{EfficientNet+Albert} baseline, we use the OCR text on each anchor and search component as the text features that are fed into Albert. 

Overall, prediction accuracy of all methods on the four task variations are reported in Tab. \ref{tab:sim-retri}. We observe that \uibert\ outperforms both baselines on all cases by 0.85\%--9.26\%, especially by a large margin on the zero-shot tasks. 
\begin{table}[]
    \centering
    \resizebox{1.0\columnwidth}{!}{
    \begin{tabular}{lcccc}
    \toprule
     & \multicolumn{2}{c}{Fine-tune} & \multicolumn{2}{c}{Zero-shot} \\
    \cmidrule(lr){2-3} \cmidrule(lr){4-5} 
    Model & Rico data & Web data & Rico data & Web data \\
    \midrule
    EfficientNet+Albert & 86.32 & 60.10 & 38.95 & 19.90 \\
    ActionBert & 85.38 & 62.85 & 30.42 & 24.80  \\
    \uibert\ & \textbf{87.90} & \textbf{63.70} & \textbf{45.53} & \textbf{34.06}  \\
    \bottomrule
    \end{tabular}
    }
    \caption{Prediction accuracy (\%) on four variations of the similar UI component retrieval task. On average, the model chooses one from 10 and 35 candidates in the Rico and web data respectively. \label{tab:sim-retri}}
\end{table}

\begin{table}[t]
    \centering
    \resizebox{\columnwidth}{!}{
    \begin{tabular}{p{1.7cm}cc|cc|c}
    \toprule
    & \multicolumn{2}{c}{Img-VH sync} & \multicolumn{2}{c}{App type cls}  & Ref exp\\
    \cmidrule(lr){2-3} \cmidrule(lr){4-5} \cmidrule(lr){6-6}
    Model & Acc. (\%) & F1 & Acc. (\%) & F1 & Acc. (\%) \\
    \midrule
    EfficientNet & & & & & \\
     +Albert  & 78.20 & 0.7500 & 68.48 & 0.6548 & 87.80\\
    ActionBert\footnote{App type cls results are different from that reported in \cite{he2021ab}, which only used a subset (43.5k out of 72k) of Rico data.}  & - & - & 72.60 & 0.6989 & 88.38 \\ 
    \uibert\ & \textbf{79.07} & \textbf{0.7706} & \textbf{72.98} & \textbf{0.7020} & \textbf{90.81}\\
    \bottomrule
    \end{tabular}
    }
    \caption{Prediction results on three downstream tasks (from left to right): image-VH sync prediction, app type classification, and referring expression component retrieval. F1 denotes Macro-F1.\label{tab:result-sync-vis-app}}
\end{table}

\begin{table}[t]
    \centering
    \resizebox{\columnwidth}{!}{
    \begin{tabular}{lcccc}
    \toprule
     & \multicolumn{2}{c}{Icon-32} & \multicolumn{2}{c}{Icon-77} \\
    \cmidrule(lr){2-3} \cmidrule(lr){4-5} 
    Model & Acc. (\%) & Macro-F1 & Acc. (\%) & Macro-F1 \\
    \midrule
    EfficientNet+Albert & 97.57 & 0.8772 & 92.52 & 0.6567 \\
    ActionBert & 97.42 & 0.8742 & 91.60 & 0.6376  \\
    \uibert\ & \textbf{97.65} & \textbf{0.8786} & \textbf{92.57} & \textbf{0.6608}  \\
    \bottomrule
    \end{tabular}
    }
    \caption{Prediction results on two icon classification tasks. \label{tab:icon-cls}}
\end{table}

\subsection{Referring Expression Component Retrieval}
Given a referring expression and a UI image, the goal of this task is to retrieve the component that the expression refers to from a set of UI components detected on the screen (Fig. \ref{fig:task-vis}(b)). This task has a practical use for voice-control systems \cite{wichers2018resolving}. Our dataset of this task is based on UIs in Rico as well. The referring expressions are collected by crowdsourcing. 
On average, the model is required to choose from 20 UI component candidates for each expression. The train, dev, and test sets respectively contain 16.9k, 2.1k and 1.8k UI componenets with their referring expressions. 

To apply \uibert\ to this task, we treat the referring expression as an OCR component and UI component candidates as IMG components that \uibert\ takes as input. Dot products of the output embedding of the expression and the output embeddings of the candidate components are computed as their similarity scores to select the referred candidate. The prediction results are shown in Tab. \ref{tab:result-sync-vis-app}. \uibert\ achieves the best accuracy 90.81\%, which outperforms ActionBert by 2.43\%. 

\subsection{Image-VH Sync Prediction\label{sec:img-sync-cls}}
View hierarchies can be noisy when they are unsynchronized with screenshots (\cite{li2020mapping}). This task takes the UI screen with its VH as input and outputs whether the VH matches the screen. It can serve as an important pre-processing step to filter out the problematic UIs. We use the RicoSCA (\cite{li2020mapping}) that have 25k synchronized and 47k unsynchronized UIs and split them into train, dev, and test sets by a ratio of 8:1:1.

We use the UI embedding of the CLS component followed by a one-layer projection to predict whether the image and view hierarchy of an UI are synced. Tab. \ref{tab:result-sync-vis-app} shows that \uibert\ outperforms the baseline and achieves 79.07\% accuracy and 77.06\% macro-F1. 

\subsection{App Type Classification\label{sec:app-cls}} 
This task aims to predict the type of an app (e.g. music, finance) of a UI. We use all the 72k unique UIs in Rico across a total of 27 app types and split them in the ratio of 8:1:1 for train, dev, and test. This task can help filter the malicious apps that have incorrect app types.

For this task, we also use a one-layer projection layer to project the \uibert\ embeddings to one of the app types. We experiment using the output of CLS component and a concatenation of the embeddings of all the UI components. The preliminary experiments show that the latter yields better results. As shown in Tab. \ref{tab:result-sync-vis-app}, \uibert\ outperforms the \emph{EfficientNet+Albert} baseline by 4.50\% accuracy and 4.72\% Macro-F1, showing the gain from the attention mechanisms of the Transformer block and from pretraining. 

\subsection{Icon Classification\label{sec:icon-cls}} 
This task aims to identify the types of icons (e.g. menu, backward, search), which is useful for applications like screen readers. We use Rico data with human-labelled icon types for every VH leaf node in two levels of granularity: 32 and 77 classes \cite{he2021ab}. 
To predict the types of an icon component, we concatenate the UI embeddings of the icon's corresponding IMG and VH components and feed them into a fully connected layer. As shown in Tab. \ref{tab:icon-cls}, \uibert\ consistently outperforms baselines in both accuracy and F1 score. 
\section{Conclusion}
We propose \uibert, a transformer-based model to learn multimodal UI representations via novel pretraining tasks. The model is evaluated on nine UI related downstream tasks and achieves the best performance across all. Visualization of UI embeddings and content descriptions generated by the pretrained model further demonstrated the efficacy of our approach. We hope our work facilitates the model development towards generic UI understanding.

\section*{Acknowledgments}

The authors thank Maria Wang,
Gabriel Schubiner, Lijuan Liu, and Nevan Wichers for their guidance and help on dataset creation and processing; James
Stout and Pranav Khaitan for
advice, guidance and encouragement; all the anonymous reviewers for reviewing the manuscript and
providing valuable feedback.

\bibliographystyle{named}
\bibliography{ijcai21}

\begin{thebibliography}{}

\bibitem[\protect\citeauthoryear{Chen \bgroup \em et al.\egroup
  }{2020}]{chen2020unblind}
Jieshan Chen, Chunyang Chen, Zhenchang Xing, Xiwei Xu, Liming Zhu, Guoqiang Li,
  and Jinshui Wang.
\newblock Unblind your apps: Predicting natural-language labels for mobile gui
  components by deep learning.
\newblock {\em arXiv preprint arXiv:2003.00380}, 2020.

\bibitem[\protect\citeauthoryear{Chung \bgroup \em et al.\egroup
  }{2014}]{chung2014empirical}
Junyoung Chung, Caglar Gulcehre, KyungHyun Cho, and Yoshua Bengio.
\newblock Empirical evaluation of gated recurrent neural networks on sequence
  modeling.
\newblock {\em arXiv preprint arXiv:1412.3555}, 2014.

\bibitem[\protect\citeauthoryear{Deka \bgroup \em et al.\egroup
  }{2017}]{Deka2017Rico}
Biplab Deka, Zifeng Huang, Chad Franzen, Joshua Hibschman, Daniel Afergan, Yang
  Li, Jeffrey Nichols, and Ranjitha Kumar.
\newblock Rico: A mobile app dataset for building data-driven design
  applications.
\newblock In {\em Proceedings of the 30th Annual Symposium on User Interface
  Software and Technology}, UIST '17, 2017.

\bibitem[\protect\citeauthoryear{Devlin \bgroup \em et al.\egroup
  }{2018}]{devlin2018bert}
Jacob Devlin, Ming-Wei Chang, Kenton Lee, and Kristina Toutanova.
\newblock Bert: Pre-training of deep bidirectional transformers for language
  understanding.
\newblock {\em arXiv preprint arXiv:1810.04805}, 2018.

\bibitem[\protect\citeauthoryear{Firebase}{2020}]{robodocs}
Firebase.
\newblock Robo app crawler documentation.
\newblock \url{https://firebase.google.com/docs/test-lab/android/robo-ux-test},
  2020.
\newblock Accessed: 2021-03-21.

\bibitem[\protect\citeauthoryear{He \bgroup \em et al.\egroup
  }{2020}]{he2021ab}
Zecheng He, Srinivas Sunkara, Xiaoxue Zang, Ying Xu, Lijuan Liu, Nevan Wichers,
  Gabriel Schubiner, Ruby Lee, and Jindong Chen.
\newblock Actionbert: Leveraging user actions for semantic understanding of
  user interfaces.
\newblock {\em arXiv preprint arXiv:2012.12350}, 2020.

\bibitem[\protect\citeauthoryear{Huang \bgroup \em et al.\egroup
  }{2019}]{huang2019swire}
Forrest Huang, John~F Canny, and Jeffrey Nichols.
\newblock Swire: Sketch-based user interface retrieval.
\newblock In {\em Proceedings of the 2019 CHI Conference on Human Factors in
  Computing Systems}, pages 1--10, 2019.

\bibitem[\protect\citeauthoryear{Jozefowicz \bgroup \em et al.\egroup
  }{2016}]{jozefowicz2016exploring}
Rafal Jozefowicz, Oriol Vinyals, Mike Schuster, Noam Shazeer, and Yonghui Wu.
\newblock Exploring the limits of language modeling.
\newblock {\em arXiv preprint arXiv:1602.02410}, 2016.

\bibitem[\protect\citeauthoryear{Kingma and Ba}{2014}]{kingma2014adam}
Diederik~P Kingma and Jimmy Ba.
\newblock Adam: A method for stochastic optimization.
\newblock {\em arXiv preprint arXiv:1412.6980}, 2014.

\bibitem[\protect\citeauthoryear{Lan \bgroup \em et al.\egroup
  }{2019}]{lan2019albert}
Zhenzhong Lan, Mingda Chen, Sebastian Goodman, Kevin Gimpel, Piyush Sharma, and
  Radu Soricut.
\newblock Albert: A lite bert for self-supervised learning of language
  representations.
\newblock In {\em International Conference on Learning Representations}, 2019.

\bibitem[\protect\citeauthoryear{Li \bgroup \em et al.\egroup
  }{2020a}]{li2020unicoder}
Gen Li, Nan Duan, Yuejian Fang, Ming Gong, Daxin Jiang, and Ming Zhou.
\newblock Unicoder-vl: A universal encoder for vision and language by
  cross-modal pre-training.
\newblock In {\em AAAI}, 2020.

\bibitem[\protect\citeauthoryear{Li \bgroup \em et al.\egroup
  }{2020b}]{li2020mapping}
Yang Li, Jiacong He, Xin Zhou, Yuan Zhang, and Jason Baldridge.
\newblock Mapping natural language instructions to mobile {UI} action
  sequences.
\newblock In {\em Proceedings of the 58th Annual Meeting of the Association for
  Computational Linguistics}, Online, July 2020. Association for Computational
  Linguistics.

\bibitem[\protect\citeauthoryear{Li \bgroup \em et al.\egroup
  }{2020c}]{li2020widget}
Yang Li, Gang Li, Luheng He, Jingjie Zheng, Hong Li, and Zhiwei Guan.
\newblock Widget captioning: Generating natural language description for mobile
  user interface elements.
\newblock In {\em Proceedings of the 2020 Conference on Empirical Methods in
  Natural Language Processing (EMNLP)}, pages 5495--5510, 2020.

\bibitem[\protect\citeauthoryear{Liu \bgroup \em et al.\egroup
  }{2018}]{liu2018learning}
Thomas~F Liu, Mark Craft, Jason Situ, Ersin Yumer, Radomir Mech, and Ranjitha
  Kumar.
\newblock Learning design semantics for mobile apps.
\newblock In {\em Proceedings of the 31st Annual ACM Symposium on User
  Interface Software and Technology}, pages 569--579, 2018.

\bibitem[\protect\citeauthoryear{Maaten and
  Hinton}{2008}]{maaten2008visualizing}
Laurens van~der Maaten and Geoffrey Hinton.
\newblock Visualizing data using t-sne.
\newblock {\em Journal of machine learning research}, 9(Nov):2579--2605, 2008.

\bibitem[\protect\citeauthoryear{MLKit}{2020}]{mlkit-ocr}
MLKit.
\newblock Recognize text in images with ml kit on android.
\newblock
  \url{https://developers.google.com/ml-kit/vision/text-recognition/android},
  2020.
\newblock Accessed: 2021-03-18.

\bibitem[\protect\citeauthoryear{Su \bgroup \em et al.\egroup
  }{2019}]{su2019vl}
Weijie Su, Xizhou Zhu, Yue Cao, Bin Li, Lewei Lu, Furu Wei, and Jifeng Dai.
\newblock Vl-bert: Pre-training of generic visual-linguistic representations.
\newblock In {\em International Conference on Learning Representations}, 2019.

\bibitem[\protect\citeauthoryear{Swearngin and
  Li}{2019}]{swearngin2019modeling}
Amanda Swearngin and Yang Li.
\newblock Modeling mobile interface tappability using crowdsourcing and deep
  learning.
\newblock In {\em Proceedings of the 2019 CHI Conference on Human Factors in
  Computing Systems}, 2019.

\bibitem[\protect\citeauthoryear{Tan and Le}{2019}]{tan2019efficientnet}
Mingxing Tan and Quoc Le.
\newblock Efficientnet: Rethinking model scaling for convolutional neural
  networks.
\newblock In {\em International Conference on Machine Learning}, 2019.

\bibitem[\protect\citeauthoryear{Vaswani \bgroup \em et al.\egroup
  }{2017}]{vaswani2017attention}
Ashish Vaswani, Noam Shazeer, Niki Parmar, Jakob Uszkoreit, Llion Jones,
  Aidan~N Gomez, {\L}ukasz Kaiser, and Illia Polosukhin.
\newblock Attention is all you need.
\newblock {\em Advances in neural information processing systems},
  30:5998--6008, 2017.

\bibitem[\protect\citeauthoryear{Wichers \bgroup \em et al.\egroup
  }{2018}]{wichers2018resolving}
Nevan Wichers, Dilek Hakkani-T{\"u}r, and Jindong Chen.
\newblock Resolving referring expressions in images with labeled elements.
\newblock In {\em 2018 IEEE Spoken Language Technology Workshop (SLT)}, pages
  800--806. IEEE, 2018.

\bibitem[\protect\citeauthoryear{Wu \bgroup \em et al.\egroup
  }{2016}]{wu2016google}
Yonghui Wu, Mike Schuster, Zhifeng Chen, Quoc~V Le, Mohammad Norouzi, Wolfgang
  Macherey, Maxim Krikun, Yuan Cao, Qin Gao, Klaus Macherey, et~al.
\newblock Google's neural machine translation system: Bridging the gap between
  human and machine translation.
\newblock {\em arXiv preprint arXiv:1609.08144}, 2016.

\end{thebibliography}

\end{document}